\documentclass[11pt,a4paper]{article}
\usepackage[hyperref]{naaclhlt2019}
\usepackage{times}
\usepackage{latexsym}
\usepackage{graphicx}
\usepackage{todonotes}
\usepackage{multirow}
\usepackage{booktabs}

\usepackage{url}

\aclfinalcopy 
\def\aclpaperid{4} 

\title{Plain English Summarization of Contracts}

\author{Laura Manor \\
  Department of Linguistics \\
  The University of Texas at Austin \\
  {\tt manor@utexas.edu} \\\And
  Junyi Jessy Li \\
  Department of Linguistics \\
  The University of Texas at Austin \\
  {\tt jessy@austin.utexas.edu} \\}

\date{}

\begin{document}
\maketitle
\begin{abstract}
 
  Unilateral contracts, such as terms of service, play a substantial role in modern digital life. However, few users read these documents before accepting the terms within, as they are too long and the language too complicated. 
  We propose the task of summarizing such legal documents in plain English, which would enable users to have a better understanding of the terms they are accepting.
  
  We propose an initial dataset of legal text snippets paired with summaries written in plain English. We verify the quality of these summaries manually and show that they involve heavy abstraction, compression, and simplification. Initial experiments show that unsupervised extractive summarization methods do not perform well on this task due to the level of abstraction and style differences. We conclude with a call for resource and technique development for simplification and style transfer for legal language.
\end{abstract}

\section{Introduction}
Although internet users accept unilateral contracts such as terms of service on a regular basis, it is well known that these users rarely read them. Nonetheless, these are binding contractual agreements. A recent study suggests that up to 98\% of users do not fully read the terms of service before accepting them \citep{obar2018biggest}. Additionally, they find that two of the top three factors users reported for not reading these documents were that they are perceived as too long (`information overload') and too complicated (`difficult to understand'). This can be seen in Table \ref{tab:pokemon}, where a section of the terms of service for a popular phone app includes a 78-word paragraph that can be distilled down to a 19-word summary.

The European Union's \citet{gdpr2018}\footnote{\url{https://eugdpr.org/}},  the  United States' \citet{plainwritingact2010}\footnote{\url{https://plainlanguage.gov/}}, and New York State's
\citet{nys1978plainenglish} show that many levels of government have recognized the need to make legal information more accessible to non-legal communities. Additionally, due to recent social movements demanding accessible and transparent policies on the use of personal data on the internet \citep{sykuta2007cori}, multiple online communities have formed that are dedicated to manually annotating various unilateral contracts.

\begin{table}[]
    \centering \small
    \begin{tabular}{p{7.3cm}} 
    \toprule
         \textbf{Original Text: }By using our Services, you are agreeing to these Terms, our Trainer Guidelines, and our Privacy Policy. If you are the parent or legal guardian of a child under the age of 13 (the “Parent”), you are agreeing to these Terms on behalf of yourself and your child(ren) who are authorized to use the Services pursuant to these Terms and in our Privacy Policy. If you don’t agree to these Terms, our Trainer Guidelines, and our Privacy Policy, do not use the Services.\\ \midrule
         \textbf{Human Summary: }By playing this game, you agree to these terms. If you're under 13 and playing, your parent/guardian agrees on your behalf. \\ \bottomrule
    \end{tabular}
    \caption{Top: an excerpt from Niantic's Pokemon GO Terms of Service. Bottom: a summary written by a community member of TLDRLegal.}
    \label{tab:pokemon}
\end{table}

We propose the task of the automatic summarization of legal documents {\em in plain English} for a non-legal audience. We hope that such a technological advancement would enable a greater number of people to enter into everyday contracts with a better understanding of what they are agreeing to. 
Automatic summarization is often used to reduce information overload, especially in the news domain \cite{nenkova2011automatic}.  Summarization has been largely missing in the legal genre, with notable exceptions of judicial judgments \cite{farzindar2004legal, hachey2006extractive} and case reports \cite{galgani2012combining}, as well as information extraction on patents \cite{tseng2007text, tang2012patentminer}. While some companies have conducted proprietary research in the summarization of contracts, this information sits behind a large pay-wall and is geared toward law professionals rather than the general public. 

In an attempt to motivate advancement in this area, we have collected  446 sets of contract sections and corresponding reference summaries which can be used as a test set for such a task.\footnote{The dataset is available at \url{ https://github.com/lauramanor/legal_summarization}} 
We have compiled these sets from two websites dedicated to explaining complicated legal documents in plain English.

Rather than attempt to summarize an entire document, these sources summarize each document at the section level. In this way, the reader can reference the more detailed text if need be.  The summaries in this dataset are reviewed for quality by the first author, who has 3 years of professional contract drafting experience.

The dataset we propose contains 446 sets of parallel text. We show the level of abstraction through the number of novel words in the reference summaries, which is significantly higher than the abstractive single-document summaries created for the shared tasks of the Document Understanding Conference (DUC) in 2002~\cite{over2007duc}, a standard dataset used for single document news summarization. Additionally, we utilize several common readability metrics to show that there is an average of a 6 year reading level difference between the original documents and the reference summaries in our legal dataset.

In initial experimentation using this dataset, we employ  popular unsupervised extractive summarization models such as TextRank~\cite{mihalcea2004textrank} and Greedy KL~\cite{haghighi2009exploring}, as well as lead baselines. We show that such methods do not perform well on this dataset when compared to the same methods on DUC 2002. These results highlight the fact that this is a very challenging task. As there is not currently a dataset in this domain large enough for supervised methods, we suggest the use of methods developed for simplification and/or style transfer.

In this paper, we begin by discussing how this task relates to the current state of text summarization and similar tasks in Section~\ref{sec:related-work}. We then introduce the novel dataset and provide details on the level of abstraction, compression, and readability in Section~\ref{sec:data}. Next, we provide results and analysis on the performance of extractive summarization baselines on our data in Section~\ref{sec:experiments}. Finally, we discuss the potential for unsupervised systems in this genre in Section~\ref{sec:discussion}.

\section{Related work} \label{sec:related-work}
Given a document, the goal of \textit{single document summarization} is to produce a shortened summary of the document that captures its main semantic content~\cite{nenkova2011automatic}. Existing research extends over several genres, including news~\cite{over2007duc,see2017get,grusky2018newsroom}, scientific writing~\cite{tac2014biomedsumm,jaidka2016overview,yasunaga2019scisummnet},  legal case reports ~\cite{galgani2012combining}, etc. A critical factor in successful summarization research is the availability of a dataset with parallel document/human-summary pairs for system evaluation. However, no such publicly available resource for summarization of contracts exists to date. We present the first dataset in this genre. 
Note that unlike other genres where human summaries paired with original documents can be found at scale, e.g., the CNN/DailyMail dataset~\cite{see2017get}, resources of this kind are yet to be curated/created for contracts. As traditional supervised summarization systems require these types of large datasets, the resources released here are intended for evaluation, rather than training. Additionally, as a first step, we restrict our initial experiments to unsupervised baselines which do not require training on large datasets.

The dataset we present summarizes contracts in {\em plain English}.
While there is no precise definition of plain English, the general philosophy is to make a text readily accessible for as many English speakers as possible.
\cite{mellinkoff2004language,tiersma2000legal}. Guidelines for plain English often suggest a preference for words with Saxon etymologies rather than a Latin/Romance etymologies, the use of short words, sentences, and paragraphs, etc.\footnote{\url{https://plainlanguage.gov/guidelines/}} \cite{tiersma2000legal,kimble2006lifting}. In this respect, the proposed task involves some level of \textit{ text simplification}, as we will discuss in Section~\ref{sec:data:readability}. However, existing resources for text simplification target literacy/reading levels~\cite{xu2015problems} or learners of English as a second language~\cite{zhu2010monolingual}. Additionally, these models are trained using Wikipedia or news articles, which are quite different from legal documents. 
These systems are trained without access to sentence-aligned parallel corpora; they only require semantically similar texts~\cite{shen2017style,yang2018unsupervised,li2018delete}. To the best of our knowledge, however, there is no existing dataset to facilitate the transfer of legal language to plain English.

\section{Data}\label{sec:data}
This section introduces a dataset compiled from two 
websites dedicated to explaining unilateral contracts in plain English: TL;DRLegal\footnote{\url{https://tldrlegal.com/}} and TOS;DR\footnote{\url{https://tosdr.org/}, CC BY-SA 3.0}. 
These websites clarify language within legal documents by providing summaries for specific sections of the original documents. 
The data was collected using Scrapy\footnote{\url{https://scrapy.org/}} and a JSON interface provided by each website's API.  
Summaries are submitted and maintained by members of the website community; neither website requires community members to be law professionals.

\subsection{TL;DRLegal}\label{sec:data:tldrlegal}
TL;DRLegal focuses mostly on software licenses, however, we only scraped documents related to specific companies rather than generic licenses (i.e. Creative Commons, etc). 
 The scraped data consists of 84 sets sourced from 9 documents: Pokemon GO Terms of Service, TLDRLegal Terms of Service, Minecraft End User Licence Agreement, YouTube Terms of Service, Android SDK License Agreement (June 2014), Google Play Game Services (May 15th, 2013), Facebook Terms of Service (Statement of Rights and Responsibilities), Dropbox Terms of Service, and Apple Website Terms of Service. 

Each set consists of a portion from the original agreement text and a summary written in plain English.  Examples of the original text and the summary are shown in Table \ref{tab:data:TLDRLegal}.

\begin{table*}[]
    \centering \small
    \begin{tabular}{p{.9in}|p{5in}} \toprule
        Source & {Facebook Terms of Service (Statement of Rights and Responsibilities) - November 15, 2013} \\
        \midrule
        Original Text & Our goal is to deliver advertising and other commercial or sponsored content that is valuable to our users and advertisers. In order to help us do that, you agree to the following: You give us permission to use your name, profile picture, content, and information in connection with commercial, sponsored, or related content (such as a brand you like) served or enhanced by us. This means, for example, that you permit a business or other entity to pay us to display your name and/or profile picture with your content or information, without any compensation to you. If you have selected a specific audience for your content or information, we will respect your choice when we use it. We do not give your content or information to advertisers without your consent. You understand that we may not always identify paid services and communications as such.\\ \midrule
        Summary  & Facebook can use any of your stuff for any reason they want without paying you, for advertising in particular.  \\ 
         
    \end{tabular}

    \begin{tabular}{p{.9in}|p{5in}}
    \toprule
        Source & {Pokemon GO Terms of Service -  July 1, 2016} \\ \midrule
        Original Text &    We may cancel, suspend, or terminate your Account and your access to your Trading Items, Virtual Money, Virtual Goods, the Content, or the Services, in our sole discretion and without prior notice, including if (a) your Account is inactive (i.e., not used or logged into) for one year; (b) you fail to comply with these Terms; (c ) we suspect fraud or misuse by you of Trading Items, Virtual Money, Virtual Goods, or other Content; (d) we suspect any other unlawful activity associated with your Account; or (e) we are acting to protect the Services, our systems, the App, any of our users, or the reputation of Niantic, TPC, or TPCI. We have no obligation or responsibility to, and will not reimburse or refund, you for any Trading Items, Virtual Money, or Virtual Goods lost due to such cancellation, suspension, or termination. You acknowledge that Niantic is not required to provide a refund for any reason, and that you will not receive money or other compensation for unused Virtual Money and Virtual Goods when your Account is closed, whether such closure was voluntary or involuntary. We have the right to offer, modify, eliminate, and/or terminate Trading Items, Virtual Money, Virtual Goods, the Content, and/or the Services, or any portion thereof, at any time, without notice or liability to you. If we discontinue the use of Virtual Money or Virtual Goods, we will provide at least 60 days’ advance notice to you by posting a notice on the Site or App or through other communications.\\ \midrule
        Summary &   If you haven't played for a year, you mess up, or we mess up, we can delete all of your virtual goods. We don't have to give them back. We might even discontinue some virtual goods entirely, but we'll give you 60 days advance notice if that happens. \\ 
         
    \end{tabular}
    
        \begin{tabular}{p{.9in}|p{5in}} \toprule
        Source & {Apple Website Terms of Service -  Nov. 20, 2009} \\ \midrule
        Original Text & Any feedback you provide at this site shall be deemed to be  non-confidential. Apple shall be free to use such information on an  unrestricted basis.\\ \midrule
        Summary &   Apple may use your feedback without restrictions (e.g. share it publicly.) \\ \bottomrule 
         
    \end{tabular}

    \caption{Examples of summary sets from TLDRLegal.}
    \label{tab:data:TLDRLegal}
\end{table*}

\subsection{TOS;DR}\label{sec:data:tosdr}
TOS;DR tends to focus on topics related to user data and privacy. We scraped 421 sets of parallel text sourced from 166 documents by 122 companies. Each set consists of a portion of an agreement text (e.g., Terms of Use, Privacy Policy, Terms of Service) and 1-3 human-written summaries. 

While the multiple references can be useful for system development and evaluation, the qualities of these summaries varied greatly. Therefore, each text was examined by 
the first author, who has three years of professional experience in contract drafting for a software company. A total of 361 sets had at least one quality summary in the set. For each, the annotator selected the most informative summary to be used in this paper. 

Of the 361 accepted summaries, more than two-thirds of them (152) are `templatic' summaries. A summary deemed templatic if it could be found in more than one summary set, either word-for-word or with just the service name changed. However, of the 152 templatic summaries which were selected as the best of their set, there were 111 unique summaries. This indicates that the templatic summaries which were selected for the final dataset are relatively unique. 

A total of 369 summaries were outright rejected for a variety of reasons, including summaries that: were a repetition of another summary for the same source snippet (291), were an exact quote of the original text (63), included opinionated language that could not be inferred from the original text (24), or only described the topic of the quote but not the content (20). We also rejected any summaries that are longer than the original texts they summarize. Annotated examples from TOS;DR can be found in Table \ref{tab:examples}. 

\begin{table*}[]
    \centering \small
    \begin{tabular}{p{.9in}|p{5in}} \toprule
        Original Text &  When you upload, submit, store, send or receive content to or through our Services, you give Google (and those we work with) a worldwide license to use, host, store, reproduce, modify, create derivative works (such as those resulting from translations, adaptations or other changes we make so that your content works better with our Services), communicate, publish, publicly perform, publicly display and distribute such content.\\ \midrule
        Summary1 (best) & The copyright license you grant is “for the limited purpose of operating, promoting, and improving” existing and new Google Services. However, please note that the license does not end if you stop using the Google services. \\  \midrule
        Summary2 & The copyright license that users grant this service is limited to the parties that make up the service's broader platform.\\
        \midrule
        Summary3 & Limited copyright license to operate and improve all Google Services \\
         
    \end{tabular}
    
        \begin{tabular}{p{0.9in}|p{5in}} \toprule
        Original Text &  We may share information with vendors, consultants, and other service providers (but not with advertisers and ad partners) who need access to such information to carry out work for us. The partner’s use of personal data will be subject to appropriate confidentiality and security measures.\\ \midrule
        Summary1 (best) & Reddit shares data with third parties \\  \midrule
        Summary2 & Third parties may be involved in operating the service\\
        \midrule
        Summary3 & {Third parties may be involved in operating the service} \\ 
        (rejected) & \\
        \bottomrule 
         
    \end{tabular}

    \caption{Examples from TOS;DR. Contract sections from TOS;DR included up to three summaries. In each case, the summaries were inspected for quality. Only the best summary was included in the analysis in this paper.}
    \label{tab:examples}
\end{table*}

\section{Analysis} \label{sec:analysis}

\subsection{Levels of abstraction and compression}\label{sec:data:metrics}

\begin{figure}[]
    \centering
    \includegraphics[width=\columnwidth]{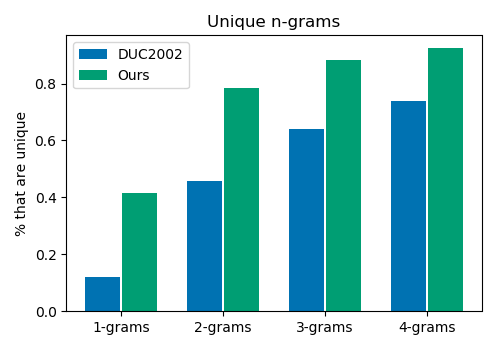}
    \caption{Unique n-grams in the reference summary, contrasting our legal dataset with DUC 2002 single document summarization data.}
    \label{fig:data:metrics:uniquen-gram}
\end{figure}
To understand the level of abstraction of the proposed dataset, we first calculate the number of n-grams that appear only in the reference summaries and not in the original texts they summarize \cite{see2017get, chen2018fast}.
As shown in Figure \ref{fig:data:metrics:uniquen-gram}, 41.4\% of words in the reference summaries did not appear in the original text. Additionally, 78.5\%, 88.4\%, and 92.3\% of 2-, 3-, and 4-grams in the reference summaries did not appear in the original text. When compared to a standard abstractive news dataset also shown in the graph (DUC 2002), the legal dataset is significantly more abstractive. 

\begin{figure}[]
    \centering
    \includegraphics[width=\columnwidth]{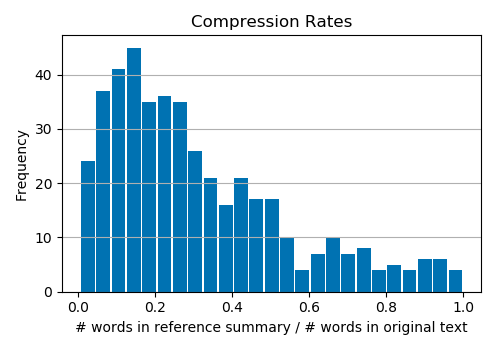}
    \caption{Ratio of words in the reference summary to words in the original text. The ratio was calculated by dividing the number of words in the reference summary by the number of words  in the original text.}
    \label{fig:data:metrics:wordratios}
\end{figure}

Furthermore, as shown in Figure \ref{fig:data:metrics:wordratios}, the dataset is very compressive, with a mean compression rate of 0.31 (std 0.23). 
The original texts have a mean of 3.6 (std 3.8) sentences per document and a mean of 105.6 (std 147.8) words per document.
The reference summaries have a mean of 1.2 (std 0.6) sentences per document, and a mean of 17.2 (std 11.8)  words per document. 

\subsection{Readability}\label{sec:data:readability}
To verify that the summaries more accessible to a wider audience, we also compare the readability of the reference summaries and the original texts.

\paragraph{Full texts} We make a comparison between the original contract sections and respective summaries using four common readability metrics. All readability metrics were implemented using Wim Muskee's readability calculator library for Python\footnote{\url{https://github.com/wimmuskee/readability-score}}. These measurements included: 
\begin{itemize}
    \item \textbf{Flesch-Kincaid formula (F-K)}: the weighted sum of the number of words in a sentence and the number of syllables per word \citep{kincaid1975derivation},
    \item \textbf{Coleman-Liau index (CL)}: the weighted sum of the number of letters per 100 words and the average number of sentences per 100 words \citep{coleman1975computer},
    \item \textbf{SMOG}: the weighted square root of the number of polysyllable words per sentence \citep{mc1969smog}, and
    \item \textbf{Automated readability index} (ARI): the weighted sum of the number of characters per word and number of words per sentence \citep{senter1967automated}. 
\end{itemize}
Though these metrics were originally formulated based on US grade levels, we have adjusted the numbers to provide the equivalent age correlated with the respective US grade level. 

We ran each measurement on the reference summaries and original texts. As shown in Table \ref{tab:data:readability:full}, the reference summaries scored lower than the original texts for each test by an average of 6 years. 

\begin{table}[]
    \centering \small
    \begin{tabular}{c|cccc|c}
               &  F-K &  C-L  &  SMOG &  ARI  &  Avg \\ \toprule 
        Ref   & 12.66 & 15.11 & 14.14  & 12.98 & 13.29\\ 
        Orig  & 20.22& 16.53  & 19.58  & 22.24  & 19.29 \\
    \end{tabular}
    \caption{Average readability scores for the reference summaries (Ref) and the original texts (Orig). Descriptions of each measurement can be found in Section \ref{sec:data:readability}.}
    \label{tab:data:readability:full}
\end{table}

\paragraph{Words}
We also seek to single out lexical difficulty, as legal text often contains vocabulary that is difficult for non-professionals. To do this, we obtain the top 50 words $W_s$ most associated with summaries and top 50 words $W_d$ most associated with the original snippets (described below) and consider the {\em differences} of ARI and F-K measures. We chose these two measures because they are a weighted sum of a word and sentential properties; as sentential information is kept the same (50 1-word ``sentences''), the {\em differences} will reflect the change in readability of the words most associated with plain English summaries/original texts.

To collect $W_s$ and $W_d$, we calculate the log odds ratio for each word, a measure used in prior work comparing summary text and original documents \cite{nye2015identification}. The log odds ratio compares the probability of a word $w$ occurring in the set of all summaries $S$ vs.\ original texts $D$:

\[ log \left ( \frac{Odds(w,S)}{Odds(w,D)} \right ) \simeq log \left ( \frac{P(w | S )}{P(w | D )} \right )\]
The list of words with the highest log odds ratios for the reference summaries ($W_s$) and original texts ($W_d$) can be found in Table \ref{tab:data:logsum}. 

We calculate the differences (in years) of ARI and F-K scores between $W_s$ and $W_d$: 
\[ARI(W_d)-ARI(W_s)=5.66\] 
\[FK(W_d)-FK(W_s)=6.12\] 
Hence, there is a $\sim$6-year reading level distinction between the two sets of words, an indication that lexical difficulty is paramount in legal text.

\begin{table}[]
    \centering \small
    
    \begin{tabular}{p{2.8in}}\toprule
        \textbf{Original Text: }arise, unless, receive, whether, example, signal, b, technology, identifier, expressly, transmit, visit, perform, search, partner, understand, conduct, server, child, support, regulation, base, similar, purchase, automatically, mobile, agent, derivative, either, commercial, reasonable, cause, functionality, advertiser, act, ii, thereof, arbitrator, attorney, modification, locate, c, individual, form, following, accordance, hereby, cookie, apps, advertisement\\ \midrule
        \textbf{Reference Summary: } fingerprint, fit, header, targeted, involve, pixel, advance, quality, track, want, stuff, even, guarantee, maintain, beacon, ban, month, prohibit, allow, defend, notification, ownership, acceptance, delete, user, prior, reason, hold, notify, govern, keep, class, change, might, illegal, old, harmless, indemnify, see, assume, deletion, waive, stop, operate, year, enforce, target, many, constitute, posting\\ \bottomrule
    \end{tabular}
    \caption{The 50 words most associated with the original text or reference summary, as measured by the log odds ratio. }
    \label{tab:data:logsum}
\end{table}

\section{Summarization baselines}\label{sec:experiments}

We present our legal dataset as a test set for contracts summarization. In this section, we report baseline performances of {\em unsupervised, extractive} methods as most recent supervised abstractive summarization methods, e.g., \citet{rush2015neural}, \citet{see2017get}, would not have enough training data in this domain. 
We chose to look at the following common baselines:
\begin{itemize}

\item\textbf{TextRank} Proposed by \citet{mihalcea2004textrank}, TextRank harnesses the PageRank algorithm to choose the sentences with the highest similarity scores to the original document.\footnote{For this paper we utilized the TextRank package from Summa NLP: \url{https://github.com/summanlp/textrank}}

\item\textbf{KLSum} An algorithm introduced by  \citep{haghighi2009exploring} which greedily selects the sentences that minimize the Kullback-Lieber (KL) divergence between the original text and proposed summary.

\item\textbf{Lead-1} A common baseline in news summarization is to select the first 1-3 sentences of the original text as the summary \cite{see2017get}. With this dataset, we include the first sentence as the summary as it is the closest to the average number of sentences per reference (1.2).

\item\textbf{Lead-K} A variation of Lead-1, this baseline selects the first k sentences until a word limit is satisfied.

\item\textbf{Random-K} This baseline selects a random sentence until a word limit is satisfied. For this baseline, the reported numbers are an average of 10 runs on the entire dataset.
\end{itemize}
    
\paragraph{Settings}
We employ lowercasing and lemmatization, as well as remove stop words and punctuation during pre-processing\footnote{NLTK was used for lemmatization and identification of stop words.}. For TextRank, KLSum, Lead-K, and Random-K, we produce summaries budgeted at the average number of words among all summaries~\cite{rush2015neural}. However, for the sentence which causes the summary to exceed the budget, we keep or discard the full sentence depending on which resulting summary is closer to the budgeted length.

\paragraph{Results}
\begin{table*}[t!]
\small
\centering
\begin{tabular}{l|c|c|c|c|c|c|c|c|c||c|c|c}
\toprule
        & \multicolumn{3}{c|}{TLDRLegal}          & \multicolumn{3}{c|}{TOS;DR} & \multicolumn{3}{c||}{Combined} & \multicolumn{3}{|c}{DUC 2002} \\ \cmidrule{2-13} 
        & R-1 & R-2 & \multicolumn{1}{c|}{R-L} & R-1 & R-2 & \multicolumn{1}{c|}{R-L} & R-1 & R-2  & \multicolumn{1}{c||}{R-L} & R-1 & R-2  & \multicolumn{1}{c}{R-L} \\ \midrule
TextRank 
 & {\bf 25.60} & 8.05 & {\bf 18.62}  & 23.88 & 6.96 & 16.96  & 24.03 & 7.16 & 17.10   & 40.94 & 18.89 & 36.70 \\ 
KLSum 
 & 24.98 & 7.84 & 18.08  & 23.25 & 6.76 & 16.67  & 23.56 & 6.94 & 16.93   & 40.06 & 16.94 & 35.85 \\ 
Lead-1 
 & 23.09 & {\bf 8.23} & 17.10  & 24.05 & 7.30 & 17.22  & 23.87 & 7.47 & 17.19   & 29.66 & 13.76 & 19.46 \\ 
Lead-K
 & 24.04 & 8.14 & 17.46  & {\bf 24.47} & {\bf 7.40} & {\bf 17.66}  & \textbf{24.38} & {\bf 7.52} & {\bf 17.63}   & \textbf{43.57} & {\bf 21.69} & {\bf 39.49} \\ 
 Random-K
  & 21.94 & 6.19 & 15.84  & 22.39 & 6.17 & 16.01  & 22.32 & 6.33 & 16.09 & 35.75 & 14.12 & 31.91 \\ \bottomrule

\end{tabular}
\caption{Performance for each dataset on the baselines was measured using Rouge-1, Rouge-2, and Rouge-L.}
\label{tab:textrank}
\end{table*}

To gain a quantitative understanding of the baseline results, we employed ROUGE~\cite{lin2004rouge}. ROUGE is a standard metric used for evaluating summaries based on the lexical overlap between a generated summary and gold/reference summaries. The ROUGE scores for the unsupervised summarization baselines found in this paper can be found in Table~\ref{tab:textrank}. 

In the same table, we also tabulate ROUGE scores of the same baselines run on DUC 2002~\cite{over2007duc}, 894 documents with summary lengths of 100 words, following the same settings. Note that our performance is a bit different from reported numbers in \citet{mihalcea2004textrank}, as we performed different pre-processing and the summary lengths were not processed in the same way.

Crucially, ROUGE scores are much higher on DUC 2002 than on our legal dataset. We speculate that this is due to the highly abstractive nature of this data, in addition to the divergent styles of the summaries and original texts.

In general, Lead-K performed best on both TOS;DR and DUC 2002.  The performance gap between TextRank and Lead-K is much larger on DUC 2002 than on our dataset. On the legal datasets, TextRank outperformed Lead-K on TLDRLegal and is very close to the performance of Lead-K on TOS;DR.
Additionally, Random-K performed only about 2 ROUGE points lower than Lead-K on our dataset, while it scored almost 8 points lower on the DUC 2002 dataset. We attribute this to the structure of the original text; 
news articles (i.e. DUC 2002) follow the inverse pyramid structure where the first few sentences give an overview of the story, and the rest of the article content is diverse. In contracts, the sentences in each section are more similar to each other lexically.

\paragraph{Qualitative Analysis}
We examined some of the results of the unsupervised extractive techniques to get a better understanding of what methods might improve the results. Select examples can be found in Table~\ref{tab:summarization:results}. 

As shown by example (1), the extractive systems performed well when the reference summaries were either an extract or a compressed version of the original text. However, examples (2-4) show various ways the extractive systems were not able to perform well. 

In (2), the extractive systems were able to select an appropriate sentence, but the sentence is much more complex than the reference summary. Utilizing text simplification techniques may help in these circumstances. 

In (3), we see that the reference summary is much better able to abstract over a larger portion of the original text than the selected sentences. (3a) shows that by having much shorter sentences, the reference summary is able to cover more of the original text. (3b) is able to restate 651-word original text in 11 words. 

Finally, in (4), the sentences from the original text are extremely long, and thus the automated summaries, while only having one sentence, are 711 and 136 words respectively. Here, we also see that the reference summaries have a much different style than the original text.

\begin{table*}[]
    \centering \small
    \setlength\tabcolsep{2.5pt}
    
     \begin{tabular}{p{.5cm}p{2.1cm}|p{12.9cm}} \toprule
            \multirow{4}{*}{(1a)} & 
            \scriptsize{Reference Summary}  & librarything will not sell or give personally identifiable information to any third party. \\ \cline{2-3}
            &  \scriptsize{TextRank, Lead-K} &  no sale of personal information. librarything will not sell or give personally identifiable information to any third party. \\ \cline{2-3}
            & \scriptsize{KLSum} &  this would be evil and we are not evil.  \\ 
        \end{tabular}
    \vspace{.1pt}
    
     \begin{tabular}{p{.5cm}p{2.1cm}|p{12.9cm}} \midrule
            \multirow{3}{*}{(1b)} & 
            \scriptsize{Reference Summary}  & you are responsible for maintaining the security of your account and for the activities on your account \\ \cline{2-3}
            & \scriptsize{TextRank,  {KLSum}, Lead-K} &  you are responsible for maintaining the confidentiality of your password and account if any and are fully responsible for any and all activities that occur under your password or account \\
        \end{tabular}
    \vspace{.1pt}
    
     \begin{tabular}{p{.5cm}p{2.1cm}|p{12.9cm}} \midrule
        \multirow{3}{*}{(2a)} & 
        \scriptsize{Reference Summary}  & if you offer suggestions to the service they become the owner of the ideas that you give them \\ \cline{2-3}
        & \scriptsize{TextRank,  {KLSum}, Lead-K}  &  if you provide a submission whether by email or otherwise you agree that it is non confidential unless couchsurfing states otherwise in writing and shall become the sole property of couchsurfing  \\ 
        \end{tabular}
    \vspace{.1pt}      
    
     \begin{tabular}{p{.5cm}p{2.1cm}|p{12.9cm}} \midrule
        \multirow{3}{*}{(2b)} & 
        \scriptsize{Reference Summary}  & when the service wants to change its terms users are notified a month or more in advance. \\ \cline{2-3}
        & \scriptsize{TextRank} & in this case you will be notified by e mail of any amendment to this agreement made by valve within 60 sixty days before the entry into force of the said amendment.   \\
        \end{tabular}
    \vspace{.1pt}
        
     \begin{tabular}{p{.5cm}p{2.1cm}|p{12.9cm}} \midrule
        \multirow{3}{*}{(2c)} & 
        \scriptsize{Reference Summary}  & you cannot delete your account for this service. \\ \cline{2-3}
        &  \scriptsize{TextRank,  {KLSum}, Lead-K}  & please note that we have no obligation to delete any of stories favorites or comments listed in your profile or otherwise remove their association with your profile or username.  \\
        \end{tabular}
    \vspace{.1pt}
    
     \begin{tabular}{p{.5cm}p{2.1cm}|p{12.9cm}} \midrule
        \multirow{9}{*}{(3a)} & 
        \scriptsize{Original Text} & by using our services you are agreeing to these terms our trainer guidelines and our privacy policy. if you are the parent or legal guardian of a child under the age of 13 the parent you are agreeing to these terms on behalf of yourself and your child ren who are authorized to use the services pursuant to these terms and in our privacy policy. if you don t agree to these terms our trainer guidelines and our privacy policy do not use the services. \\ \cline{2-3}
        &\scriptsize{Reference Summary}  & if you don t agree to these terms our trainer guidelines and our privacy policy do not use the services. \\ \cline{2-3}
        & \scriptsize{TextRank} & by playing this game you agree to these terms. if you re under 13 and playing your parent guardian agrees on your behalf.  \\ \cline{2-3}
        & \scriptsize{{KLSum}, Lead-K} & by using our services you are agreeing to these terms our trainer guidelines and our privacy policy. \\ 
        \end{tabular}
    \vspace{.1pt}        
        
     \begin{tabular}{p{.5cm}p{2.1cm}|p{12.9cm}} \midrule
        \multirow{15}{*}{(3b)} & 
        \scriptsize{Original Text} & subject to your compliance with these terms niantic grants you a limited nonexclusive nontransferable non sublicensable license to download and install a copy of the app on a mobile device and to run such copy of the app solely for your own personal noncommercial purposes. [...] by using the app you represent and warrant that i you are not located in a country that is subject to a u s government embargo or that has been designated by the u s government as a terrorist supporting country and ii you are not listed on any u s government list of prohibited or restricted parties.  \\ \cline{2-3}
        & \scriptsize{Reference Summary}  &  \\ \cline{2-3}
        & \scriptsize{TextRank} & in the event of any third party claim that the app or your possession and use of the app infringes that third party s intellectual property rights niantic will be solely responsible for the investigation defense settlement and discharge of any such intellectual property infringement claim to the extent required by these terms.   \\ \cline{2-3}
        & \scriptsize{KLSum} & if you accessed or downloaded the app from any app store or distribution platform like the apple store google play or amazon appstore each an app provider then you acknowledge and agree that these terms are concluded between you and niantic and not with app provider and that as between us and the app provider niantic is solely responsible for the app. \\ 
        \end{tabular}
    \vspace{.1pt}
        
     \begin{tabular}{p{.5cm}p{2.1cm}|p{12.9cm}} \midrule
        \multirow{7}{*}{(4a)} & 
            \scriptsize{Reference Summary}  & don t be a jerk. don t hack or cheat. we don t have to ban you but we can. we ll also cooperate with law enforcement. \\ \cline{2-3}
            & \scriptsize{KLSum} &  by way of example and not as a limitation you agree that when using the services and content you will not defame abuse harass harm stalk threaten or otherwise violate the legal rights including the rights of privacy and publicity of others [...]  lease the app or your account collect or store any personally identifiable information from the services from other users of the services without their express permission violate any applicable law or regulation or enable any other individual to do any of the foregoing. \\ 
    \end{tabular}
    \vspace{.1pt}    
    
     \begin{tabular}{p{.5cm}p{2.1cm}|p{12.9cm}}\midrule
        \multirow{9}{*}{(4b)} & 
            \scriptsize{Reference Summary}  & don t blame google. \\ \cline{2-3}
            & \scriptsize{TextRank,  {KLSum}, Lead-K}   &  the indemnification provision in section 9 of the api tos is deleted in its entirety and replaced with the following you agree to hold harmless and indemnify google and its subsidiaries affiliates officers agents and employees or partners from and against any third party claim arising from or in any way related to your misuse of google play game services your violation of these terms or any third party s misuse of google play game services or actions that would constitute a violation of these terms provided that you enabled such third party to access the apis or failed to take reasonable steps to prevent such third party from accessing the apis including any liability or expense arising from all claims losses damages actual and consequential suits judgments litigation costs and attorneys fees of every kind and nature.  \\ \bottomrule
        \end{tabular}

    \caption{Examples of reference summaries and results from various extractive summarization techniques. The text shown here has been pre-processed. To conserve space, original texts were excluded from most examples. }
    \label{tab:summarization:results}
\end{table*}

\section{Discussion} \label{sec:discussion}
Our preliminary experiments and analysis show that summarizing legal contracts in plain English is challenging, and point to the potential usefulness of a simplification or style transfer system in the summarization pipeline. Yet this is challenging. First, there may be a substantial domain gap between legal documents and texts that existing simplification systems are trained on (e.g., Wikipedia, news). Second, popular supervised approaches such as treating sentence simplification as monolingual machine translation~\cite{specia2010translating,zhu2010monolingual,woodsend2011learning,xu2016optimizing,zhang2017sentence} would be difficult to apply due to the lack of sentence-aligned parallel corpora. Possible directions include unsupervised lexical simplification utilizing distributed representations of words~\cite{glavas2015simplifying,paetzold2016unsupervised},  unsupervised sentence simplification using rich semantic structure ~\cite{narayan2016unsupervised}, or unsupervised style transfer techniques~\cite{shen2017style,yang2018unsupervised,li2018delete}. However, there is not currently a dataset in this domain large enough for unsupervised methods, nor corpora unaligned but comparable in semantics across legal and plain English, which we see as a call for future research.

\section{Conclusion} \label{sec:conclusion}
In this paper, we propose the task of summarizing legal documents in plain English and present an initial evaluation dataset for this task. We gather our dataset from online sources dedicated to explaining sections of contracts in plain English and manually verify the quality of the summaries. We show that our dataset is highly abstractive and that the summaries are much simpler to read. This task is challenging, as popular unsupervised extractive summarization methods do not perform well on this dataset and, as discussed in section \ref{sec:discussion}, current methods that address the change in register are mostly supervised as well.  We call for the development of resources for unsupervised simplification and style transfer in this domain.

\section*{Acknowledgments}
We would like to personally thank Katrin Erk for her help in the conceptualization of this project. 
Additional thanks to May Helena Plumb, Barea Sinno, and David Beavers for their aid in the revision process. 
We are grateful for the anonymous reviewers and for the TLDRLegal and TOS;DR communities and their pursuit of transparency. 


\bibliography{naaclhlt2019}

\begin{thebibliography}{42}
\expandafter\ifx\csname natexlab\endcsname\relax\def\natexlab#1{#1}\fi

\bibitem[{Chen and Bansal(2018)}]{chen2018fast}
Yen-Chun Chen and Mohit Bansal. 2018.
\newblock Fast abstractive summarization with reinforce-selected sentence
  rewriting.
\newblock In \emph{Proceedings of the 56th Annual Meeting of the Association
  for Computational Linguistics (Vol 1: Long Papers)}, pages 675--686.

\bibitem[{Coleman and Liau(1975)}]{coleman1975computer}
Meri Coleman and Ta~Lin Liau. 1975.
\newblock A computer readability formula designed for machine scoring.
\newblock \emph{Journal of Applied Psychology}, 60(2):283.

\bibitem[{Farzindar and Lapalme(2004)}]{farzindar2004legal}
Atefeh Farzindar and Guy Lapalme. 2004.
\newblock Legal text summarization by exploration of the thematic structure and
  argumentative roles.
\newblock \emph{Text Summarization Branches Out}.

\bibitem[{Galgani et~al.(2012)Galgani, Compton, and
  Hoffmann}]{galgani2012combining}
Filippo Galgani, Paul Compton, and Achim Hoffmann. 2012.
\newblock Combining different summarization techniques for legal text.
\newblock In \emph{Proceedings of the Workshop on Innovative Hybrid Approaches
  to the Processing of Textual Data}, pages 115--123.

\bibitem[{{General Data Protection Regulation}(2018)}]{gdpr2018}
{General Data Protection Regulation}. 2018.
\newblock Regulation on the protection of natural persons with regard to the
  processing of personal data and on the free movement of such data, and
  repealing directive 95/46/ec (data protection directive).
\newblock L119, 4 May 2016, pages 1–88.

\bibitem[{Glava{\v{s}} and {\v{S}}tajner(2015)}]{glavas2015simplifying}
Goran Glava{\v{s}} and Sanja {\v{S}}tajner. 2015.
\newblock Simplifying lexical simplification: Do we need simplified corpora?
\newblock In \emph{Proceedings of the 53rd Annual Meeting of the Association
  for Computational Linguistics and the 7th International Joint Conference on
  Natural Language Processing (Vol 2: Short Papers)}, pages 63--68.

\bibitem[{Grusky et~al.(2018)Grusky, Naaman, and Artzi}]{grusky2018newsroom}
Max Grusky, Mor Naaman, and Yoav Artzi. 2018.
\newblock Newsroom: A dataset of 1.3 million summaries with diverse extractive
  strategies.
\newblock In \emph{Proceedings of the 2018 Conference of the North American
  Chapter of the Association for Computational Linguistics: Human Language
  Technologies, Volume 1 (Long Papers)}, pages 708--719.

\bibitem[{Hachey and Grover(2006)}]{hachey2006extractive}
Ben Hachey and Claire Grover. 2006.
\newblock Extractive summarisation of legal texts.
\newblock \emph{Artificial Intelligence and Law}, 14(4):305--345.

\bibitem[{Haghighi and Vanderwende(2009)}]{haghighi2009exploring}
Aria Haghighi and Lucy Vanderwende. 2009.
\newblock Exploring content models for multi-document summarization.
\newblock In \emph{Proceedings of Human Language Technologies: The 2009 Annual
  Conference of the North American Chapter of the Association for Computational
  Linguistics}, pages 362--370.

\bibitem[{Jaidka et~al.(2016)Jaidka, Chandrasekaran, Rustagi, and
  Kan}]{jaidka2016overview}
Kokil Jaidka, Muthu~Kumar Chandrasekaran, Sajal Rustagi, and Min-Yen Kan. 2016.
\newblock Overview of the cl-scisumm 2016 shared task.
\newblock In \emph{Proceedings of the Joint Workshop on Bibliometric-enhanced
  Information Retrieval and Natural Language Processing for Digital Libraries},
  pages 93--102.

\bibitem[{Kimble(2006)}]{kimble2006lifting}
Joseph Kimble. 2006.
\newblock \emph{Lifting the fog of legalese: essays on plain language}.
\newblock Carolina Academic Press.

\bibitem[{Kincaid et~al.(1975)Kincaid, Fishburne~Jr, Rogers, and
  Chissom}]{kincaid1975derivation}
J~Peter Kincaid, Robert~P Fishburne~Jr, Richard~L Rogers, and Brad~S Chissom.
  1975.
\newblock Derivation of new readability formulas (automated readability index,
  fog count and flesch reading ease formula) for navy enlisted personnel.
\newblock In \emph{Technical Report, Institute for Simulation and Training,
  University of Central Florida}.

\bibitem[{Li et~al.(2018)Li, Jia, He, and Liang}]{li2018delete}
Juncen Li, Robin Jia, He~He, and Percy Liang. 2018.
\newblock Delete, retrieve, generate: a simple approach to sentiment and style
  transfer.
\newblock In \emph{Proceedings of the 2018 Conference of the North American
  Chapter of the Association for Computational Linguistics: Human Language
  Technologies, Volume 1 (Long Papers)}, pages 1865--1874.

\bibitem[{Lin(2004)}]{lin2004rouge}
Chin-Yew Lin. 2004.
\newblock {ROUGE}: A package for automatic evaluation of summaries.
\newblock In \emph{Text Summarization Branches Out}.

\bibitem[{Mc~Laughlin(1969)}]{mc1969smog}
G~Harry Mc~Laughlin. 1969.
\newblock Smog grading-a new readability formula.
\newblock \emph{Journal of reading}, 12(8):639--646.

\bibitem[{Mellinkoff(2004)}]{mellinkoff2004language}
David Mellinkoff. 2004.
\newblock \emph{The language of the law}.
\newblock Wipf and Stock Publishers.

\bibitem[{Mihalcea and Tarau(2004)}]{mihalcea2004textrank}
Rada Mihalcea and Paul Tarau. 2004.
\newblock Textrank: Bringing order into text.
\newblock In \emph{Proceedings of the 2004 conference on empirical methods in
  natural language processing}.

\bibitem[{Narayan and Gardent(2016)}]{narayan2016unsupervised}
Shashi Narayan and Claire Gardent. 2016.
\newblock Unsupervised sentence simplification using deep semantics.
\newblock In \emph{The 9th International Natural Language Generation
  conference}, pages 111--120.

\bibitem[{Nenkova et~al.(2011)Nenkova, McKeown et~al.}]{nenkova2011automatic}
Ani Nenkova, Kathleen McKeown, et~al. 2011.
\newblock Automatic summarization.
\newblock \emph{Foundations and Trends{\textregistered} in Information
  Retrieval}, 5(2--3):103--233.

\bibitem[{Nye and Nenkova(2015)}]{nye2015identification}
Benjamin Nye and Ani Nenkova. 2015.
\newblock Identification and characterization of newsworthy verbs in world
  news.
\newblock In \emph{Proceedings of the 2015 Conference of the North American
  Chapter of the Association for Computational Linguistics: Human Language
  Technologies}, pages 1440--1445.

\bibitem[{Obar and Oeldorf-Hirsch(2018)}]{obar2018biggest}
Jonathan~A Obar and Anne Oeldorf-Hirsch. 2018.
\newblock The biggest lie on the internet: Ignoring the privacy policies and
  terms of service policies of social networking services.
\newblock \emph{Information, Communication \& Society}, pages 1--20.

\bibitem[{Over et~al.(2007)Over, Dang, and Harman}]{over2007duc}
Paul Over, Hoa Dang, and Donna Harman. 2007.
\newblock {DUC} in context.
\newblock \emph{Information Processing \& Management}, 43(6):1506--1520.

\bibitem[{Paetzold and Specia(2016)}]{paetzold2016unsupervised}
Gustavo~H. Paetzold and Lucia Specia. 2016.
\newblock Unsupervised lexical simplification for non-native speakers.
\newblock In \emph{Proceedings of the 13th Association for the Advancement of
  Artificial Intelligence Conference on Artificial Intelligence}, pages
  3761--3767.

\bibitem[{{Plain English law}(1978)}]{nys1978plainenglish}
{Plain English law}. 1978.
\newblock Title 7: Requirements for use of plain language in consumer
  transactions.
\newblock The Laws Of New York Consolidated Laws. General Obligations. Article
  5: Creation, Definition And Enforcement Of Contractual Obligations.

\bibitem[{{Plain Writing Act}(2010)}]{plainwritingact2010}
{Plain Writing Act}. 2010.
\newblock An act to enhance citizen access to government information and
  services by establishing that government documents issued to the public must
  be written clearly, and for other purposes.
\newblock House of Representatives 946; Public Law No. 111-274; 124 Statues at
  Large 2861.

\bibitem[{Rush et~al.(2015)Rush, Chopra, and Weston}]{rush2015neural}
Alexander~M. Rush, Sumit Chopra, and Jason Weston. 2015.
\newblock A neural attention model for abstractive sentence summarization.
\newblock In \emph{Proceedings of the 2015 Conference on Empirical Methods in
  Natural Language Processing}, pages 379--389.

\bibitem[{See et~al.(2017)See, Liu, and Manning}]{see2017get}
Abigail See, Peter~J Liu, and Christopher~D Manning. 2017.
\newblock Get to the point: Summarization with pointer-generator networks.
\newblock In \emph{Proceedings of the 55th Annual Meeting of the Association
  for Computational Linguistics (Vol 1: Long Papers)}, pages 1073--1083.

\bibitem[{Senter and Smith(1967)}]{senter1967automated}
RJ~Senter and Edgar~A Smith. 1967.
\newblock Automated readability index.
\newblock Technical report, Cincinnati Univ OH.

\bibitem[{Shen et~al.(2017)Shen, Lei, Barzilay, and Jaakkola}]{shen2017style}
Tianxiao Shen, Tao Lei, Regina Barzilay, and Tommi Jaakkola. 2017.
\newblock Style transfer from non-parallel text by cross-alignment.
\newblock In \emph{Advances in neural information processing systems}, pages
  6830--6841.

\bibitem[{Specia(2010)}]{specia2010translating}
Lucia Specia. 2010.
\newblock Translating from complex to simplified sentences.
\newblock In \emph{Proceedings of the International Conference on Computational
  Processing of the Portuguese Language}, pages 30--39.

\bibitem[{Sykuta et~al.(2007)Sykuta, Klein, and Cutts}]{sykuta2007cori}
Michael~E Sykuta, Peter~G Klein, and James Cutts. 2007.
\newblock Cori k-base: Data overview.

\bibitem[{TAC(2014)}]{tac2014biomedsumm}
TAC. 2014.
\newblock In \emph{https://tac.nist.gov/2014/BiomedSumm/}.

\bibitem[{Tang et~al.(2012)Tang, Wang, Yang, Hu, Zhao, Yan, Gao, Huang, Xu, Li
  et~al.}]{tang2012patentminer}
Jie Tang, Bo~Wang, Yang Yang, Po~Hu, Yanting Zhao, Xinyu Yan, Bo~Gao, Minlie
  Huang, Peng Xu, Weichang Li, et~al. 2012.
\newblock Patentminer: topic-driven patent analysis and mining.
\newblock In \emph{Proceedings of the 18th Internationasl Conference on
  Knowledge Discovery and Data Mining}, pages 1366--1374.

\bibitem[{Tiersma(2000)}]{tiersma2000legal}
Peter~M Tiersma. 2000.
\newblock \emph{Legal language}.
\newblock University of Chicago Press.

\bibitem[{Tseng et~al.(2007)Tseng, Lin, and Lin}]{tseng2007text}
Yuen-Hsien Tseng, Chi-Jen Lin, and Yu-I Lin. 2007.
\newblock Text mining techniques for patent analysis.
\newblock \emph{Information Processing \& Management}, 43(5):1216--1247.

\bibitem[{Woodsend and Lapata(2011)}]{woodsend2011learning}
Kristian Woodsend and Mirella Lapata. 2011.
\newblock Learning to simplify sentences with quasi-synchronous grammar and
  integer programming.
\newblock In \emph{Proceedings of the conference on empirical methods in
  natural language processing}, pages 409--420.

\bibitem[{Xu et~al.(2015)Xu, Callison-Burch, and Napoles}]{xu2015problems}
Wei Xu, Chris Callison-Burch, and Courtney Napoles. 2015.
\newblock Problems in current text simplification research: New data can help.
\newblock \emph{Transactions of the Association for Computational Linguistics},
  3:283--297.

\bibitem[{Xu et~al.(2016)Xu, Napoles, Pavlick, Chen, and
  Callison-Burch}]{xu2016optimizing}
Wei Xu, Courtney Napoles, Ellie Pavlick, Quanze Chen, and Chris Callison-Burch.
  2016.
\newblock Optimizing statistical machine translation for text simplification.
\newblock \emph{Transactions of the Association for Computational Linguistics},
  4:401--415.

\bibitem[{Yang et~al.(2018)Yang, Hu, Dyer, Xing, and
  Berg-Kirkpatrick}]{yang2018unsupervised}
Zichao Yang, Zhiting Hu, Chris Dyer, Eric~P Xing, and Taylor Berg-Kirkpatrick.
  2018.
\newblock Unsupervised text style transfer using language models as
  discriminators.
\newblock In \emph{Advances in Neural Information Processing Systems}, pages
  7298--7309.

\bibitem[{Yasunaga et~al.(2019)Yasunaga, Kasai, Zhang, Dan, and
  Radev}]{yasunaga2019scisummnet}
Michihiro Yasunaga, Jungo Kasai, Rui Zhang, Alexander R Fabbri Irene~Li Dan,
  and Friedman Dragomir~R Radev. 2019.
\newblock Scisummnet: A large annotated corpus and content-impact models for
  scientific paper summarization with citation networks.
\newblock In \emph{Proceedings of the 13th Association for the Advancement of
  Artificial Intelligence Conference on Artificial Intelligence}.

\bibitem[{Zhang and Lapata(2017)}]{zhang2017sentence}
Xingxing Zhang and Mirella Lapata. 2017.
\newblock Sentence simplification with deep reinforcement learning.
\newblock In \emph{Proceedings of the 2017 Conference on Empirical Methods in
  Natural Language Processing}, pages 584--594.

\bibitem[{Zhu et~al.(2010)Zhu, Bernhard, and Gurevych}]{zhu2010monolingual}
Zhemin Zhu, Delphine Bernhard, and Iryna Gurevych. 2010.
\newblock A monolingual tree-based translation model for sentence
  simplification.
\newblock In \emph{Proceedings of the 23rd international conference on
  computational linguistics}, pages 1353--1361.

\end{thebibliography}
\bibliographystyle{acl_natbib}

\end{document}